\documentclass{article}

\PassOptionsToPackage{numbers, compress}{natbib}

\usepackage[pdftex]{graphicx}

\usepackage[preprint]{ARLET_2024}

\usepackage[utf8]{inputenc} 
\usepackage[T1]{fontenc}    
\usepackage{hyperref}       
\usepackage{url}            
\usepackage{booktabs}       
\usepackage{amsfonts}       
\usepackage{nicefrac}       
\usepackage{microtype}      
\usepackage{xcolor}         
\usepackage{adjustbox}
\usepackage{subcaption}
\usepackage{algorithm}
\usepackage{algpseudocode}
\usepackage{amssymb}
\usepackage{pifont}
\usepackage{multirow}
\usepackage{mathtools}
\usepackage{wrapfig}

\title{Shared-unique Features and Task-aware Prioritized Sampling on Multi-task Reinforcement Learning}

\author{Po-Shao Lin$^{1, \ast}$, \quad Jia-Fong Yeh$^{1, \ast}$, \quad Yi-Ting Chen$^{2}$, \quad Winston H. Hsu$^{2, 3}$\\
 $^{\ast}$Equal Contribution\\
 $^{1}$National Taiwan University\\
 $^{2}$National Yang Ming Chiao Tung University\\
 $^{3}$MobileDrive}

\begin{document}

\maketitle

\begin{abstract}
We observe that current state-of-the-art (SOTA) methods suffer from the performance imbalance issue when performing multi-task reinforcement learning (MTRL) tasks. While these methods may achieve impressive performance on average, they perform extremely poorly on a few tasks. To address this, we propose a new and effective method called STARS, which consists of two novel strategies: a shared-unique feature extractor and task-aware prioritized sampling. First, the shared-unique feature extractor learns both shared and task-specific features to enable better synergy of knowledge between different tasks. Second, the task-aware sampling strategy is combined with the prioritized experience replay for efficient learning on tasks with poor performance. The effectiveness and stability of our STARS are verified through experiments on the mainstream Meta-World benchmark. From the results, our STARS statistically outperforms current SOTA methods and alleviates the performance imbalance issue. Besides, we visualize the learned features to support our claims and enhance the interpretability of STARS.

\end{abstract}

\section{Introduction}
\label{sec:intro}
Multi-task reinforcement learning (MTRL) has emerged as a crucial learning paradigm. Humans have demonstrated an incredible ability to multitask; for instance, a single person can complete various everyday household tasks. Building upon this foundation, MTRL investigates whether a single robot can also possess similar multitasking capabilities. In contrast to conventional RL, where a policy is trained for each task individually, MTRL requires a policy to learn from a fixed set of tasks and validate its performance on the same set. This learning paradigm has recently garnered notable attention from the research community \cite{tanaka2003multitask, borsa2016learning, haarnoja2018soft, yang2020multi, yu2020gradient, sodhani2021multi, sun2022paco, calandriello2014sparse, wilson2007multi, vithayathil2020survey, devin2017learning, d2020sharing}.

\begin{figure}
    \centering
    \includegraphics[width=1.05\textwidth]{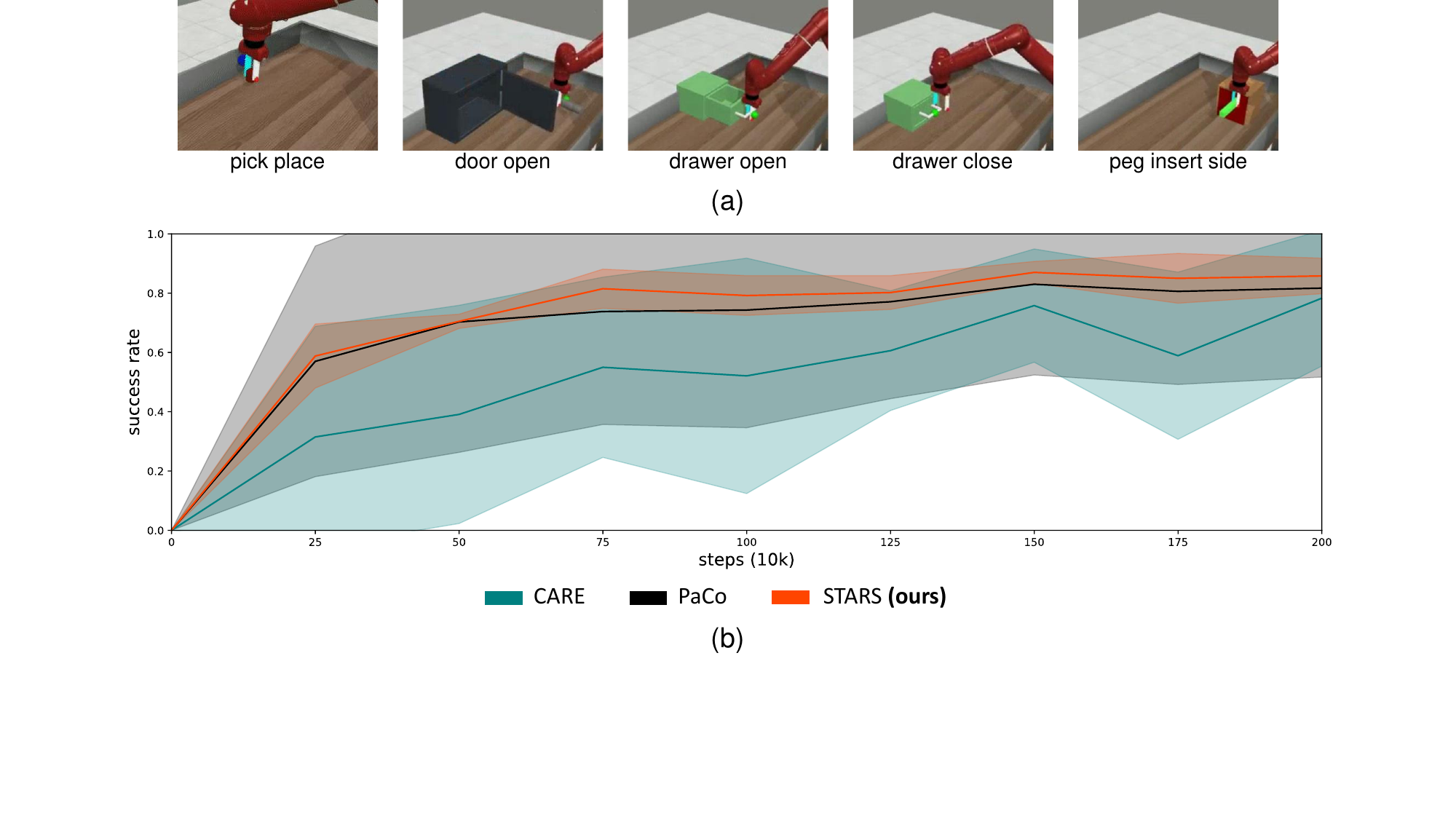}
    \vskip -0.65in
    \caption{Performance imbalance issue. (a) Selected tasks from the Meta-World benchmark \cite{yu2020meta}. Tasks like \textit{door-open} and \textit{drawer-open} share some related skills while \textit{peg-insert-side} may contains more unique skills. (b) Performance comparison between ours and previous MTRL methods. The results are averaged \textit{across tasks} and the colored area is the standard deviation (STD). The larger STD values indicate that the more severe the method suffers from the performance imbalance issue. Our method achieves the best average performance and consistently maintains a lower standard deviation.}
    \label{fig:issue}
    \vskip -0.05in
\end{figure}

Unfortunately, we observe that previous state-of-the-art (SOTA) methods \cite{yang2020multi, sodhani2021multi, sun2022paco} in MTRL suffer from the \textbf{performance imbalance issue}. We emphasize that the average performance across multiple experiment runs, the commonly reported metric in previous MTRL studies, cannot effectively reflect this serious issue. We should also consider the method performance \textit{across tasks} for a more comprehensive understanding. In Figure \ref{fig:issue}, the colored area represents the standard deviation of each method's performance \textit{across tasks}. It is evident that even though these methods achieve similar average performance across tasks, they still exhibit significant performance differences on individual tasks at the end of training, resulting in large standard deviation values (colored areas). The learning process of each method on each task is visualized in Figure \ref{fig:issue_per_task} in the Appendix.

We identify two possible reasons why previous methods suffer from the performance imbalance issue: (1) Inability to leverage both shared and unique features from tasks. Across all tasks, some may be accomplished using similar knowledge, while others may require unique knowledge, as depicted in Figure \ref{fig:issue}. Previous methods either use a unified parameter set to learn shared features from all tasks and then extract the required parts when performing respective tasks \cite{sun2022paco}, or they directly extract task-specific features using a routing network or a pretrained language metadata \cite{yang2020multi, sodhani2021multi}. However, none of these methods effectively balance exploiting shared and unique knowledge of tasks, resulting in significant performance differences across tasks, as discussed in \cite{meng2023seek}. (2) Lack of dynamic adjustment in attention between tasks based on performance differences. Previous MTRL method usually apply basic experience replay and gather the same number of samples for all tasks, which may allow some tasks' information to dominate the whole learning process.

To this end, we propose a new and effective method named \textbf{S}hare-unique features and \textbf{Ta}sk-aware \textbf{P}rioritized \textbf{S}ampling (STARS) to address the performance imbalance issue. First, STARS is equipped with a shared-unique feature extractor to achieve better knowledge synergy. This extractor retrieves knowledge that can be shared between different tasks \textbf{(shared features)} and knowledge that requires special attention in individual tasks \textbf{(unique features)}. The extracted unique features are guided by triplet loss \cite{Schroff15} to learn task-specific knowledge. Second, we develop a task-aware sampling strategy integrated with prioritized experience replay \cite{schaul2015prioritized} to dynamically adjust the number of samples requested from the replay buffer of different tasks, increasing focus on tasks with poor current learning outcomes. With these two novel designs, STARS should achieve better performance with higher stability across tasks compared to previous MTRL methods.

We evaluate the effectiveness of our STARS through experiments on the Meta-World benchmark \cite{yu2020meta}, which provides a diverse set of robot manipulation tasks. The benchmark contains two tracks, MT-10 and MT50, where the value represents the number of tasks. In the MT-10 track, our STARS statistically outperforms previous SOTA methods (6.5$\%$ improvement) and demonstrates higher stability across different tasks. We also provide detailed ablation studies and visualizations of learned unique features to validate our architecture design and support our claims. In the MT-50 track, our STARS achieves slightly better results compared to others. Enhancing knowledge synergy and optimizing training resource allocation across more tasks will be our future research direction.

\paragraph{Contributions} The key contributions of our work are summarized as follows: (1) We point out a serious performance imbalance issue from which SOTA methods in MTRL suffer. (2) We introduce STARS, a novel MTRL method that achieves the best average performance and alleviates the performance imbalance issue by leveraging a share-unique feature extractor and task-aware prioritized sampling. (3) We evaluate STARS and previous SOTA methods on the Meta-World benchmark. The results suggest that STARS statistically outperforms the compared methods in the MT-10 track. The visualizations of learned features also support our claims.

\section{Related Work}
\label{sec:related}

\paragraph{Multi-task Learning} Multi-task learning (MTL) itself is a long-standing paradigm \cite{caruana1997multitask, andreas2017modular, sener2018multi, zhang2021survey, ruder2017overview, zhang2018overview, pinto2017learning},with the primary objective of solving multiple learning tasks simultaneously. MTL methods exploit commonalities and differences across tasks, which is particularly beneficial when tasks are related, as the information gained from one task can aid in the learning of others. This leads to more robust and efficient models compared to those trained independently on each task. Besides using RL objectives (MTRL works), researchers also study MTL leveraging imitation learning (multi-task imitation learning) \cite{Singh20, Xihan22, Zhang23} or evolutionary algorithms (evolutionary multitasking) \cite{Feng19, Xu21, Osaba22}. In this work, we examine the MTRL paradigm and highlight the serious performance imbalance issue.

\paragraph{Conventional RL} Reinforcement learning (RL) is a learning paradigm where policies learn to make decisions by interacting with an environment to maximize cumulative rewards. RL has been successfully applied in various domains, including robotics \cite{Gu17, Kalashnikov18}, game playing \cite{Mnih13, Goldwaser20}, and autonomous driving \cite{Kiran22, Gu23}. RL algorithms can be categorized into three groups: Q-learning \cite{Mnih15, Hasselt16}, policy gradients \cite{Schulman15, Schulman17}, and actor-critic methods \cite{Lillicrap16, Haarnoja18}. Notably, in conventional RL, a policy is trained for a single environment or task. However, in multi-task RL, a single policy learns from a fixed set of tasks to acquire the capability to perform across those tasks. 
Our work proposes a new MTRL method that effectively address the performance imbalance issue.

\paragraph{Experience Replay} Experience replay is a technique used for storing a policy's experiences, typically in the form of state-action-reward-next state tuples, in a replay buffer. It is commonly used with off-policy RL methods, where the policies for data collection and evaluation differ. Existing experience replay research \cite{wang2016sample, zha2019experience, andrychowicz2017hindsight, Fedus20, Lu23} focuses on various features, such as maintaining a high diversity of stored experiences or giving higher priority to experiences that are currently poorly learned. However, there is a lack of experience replay techniques specifically for the MTRL setting. Consequently, previous MTRL works usually leverage basic experience replay to store experiences from each task. Our work introduces a task-aware sampling strategy with prioritized experience replay \cite{schaul2015prioritized}, enabling our method to better focus on tasks with poor performance.
\section{Preliminaries}
\label{sec:preliminary}

\paragraph{Task Statement of MTRL} In multi-task reinforcement learning (MTRL), we aim to learn a policy that performs well across multiple tasks. Let $\mathcal{T} = \{ T_{1}, T_{j}, ..., T_{N} \}, \: j = 1, 2, ..., N$ represent a set of tasks, where each task $T_{j}$ is modeled as a Markov Decision Process (MDP) $\mathcal{M}_{j} \coloneqq (\mathcal{S}_{j}, \mathcal{A}_{j}, P_{j}, R_{j}, \gamma_{j})$. Here, $\mathcal{S}_{j}$ and $\mathcal{A}_{j}$ are the state and action spaces, $P_{j}: \mathcal{S}_{j} \times \mathcal{A}_{j} \times \mathcal{S}_{j} \rightarrow [0, 1]$ is the transition probability function, $R_{j}: \mathcal{S}_{j} \times \mathcal{A}_{j} \rightarrow \mathbb{R}$ is the reward function, and $\gamma_{j}$ is the discount factor for task $T_{j}$. The objective is to learn a shared policy $\pi_{\theta}$ with parameters $\theta$ that maximizes the expected return across all tasks: 

\begin{equation}
    \max_{\theta} \sum_{j=1}^{N} \mathbb{E}_{s \sim \rho_{j}, a \sim \pi_{\theta}(\dot \mid s)} \left[ \sum_{t=0}^{\infty} \gamma_{j}^{t} R_{j}(s_{t}, a_{t}) \right],
\end{equation}

where $\rho_{j}$ is the initial state distribution for task $T_{j}$. Additionally, the task embedding or index is usually passed into the policy to indicate which task the policy is currently solving, resulting in a task-conditioned policy $\pi_{\theta}(a \mid s, T_{j})$. However, this is not a strict requirement in the general form of the MTRL task. In practice, a maximum number of environment interactions is set, and during this period, the best results obtained by each method will be used for performance comparison.

\paragraph{Soft Actor-Critic (SAC)} We use the Soft Actor-Critic (SAC) algorithm \cite{haarnoja2018soft} to train the universal policy across all tasks, following the settings from existing MTRL works \cite{sodhani2021multi, sun2022paco}. SAC is a powerful model-free, off-policy RL algorithm that aims to achieve a good balance between exploration and exploitation. Its key innovations include a policy network updated by maximizing the expected return while also considering the entropy of the policy, a soft Q-function (composed of twin Q-networks) to mitigate positive bias in the policy improvement step, and a temperature parameter to adjust the target entropy. By balancing the trade-off between exploration (entropy) and exploitation (reward), SAC achieves SOTA performance across various benchmarks.

\section{Methodology}
\label{sec:method}

\begin{figure}[!t]
    \centering
    \includegraphics[width=1\linewidth]{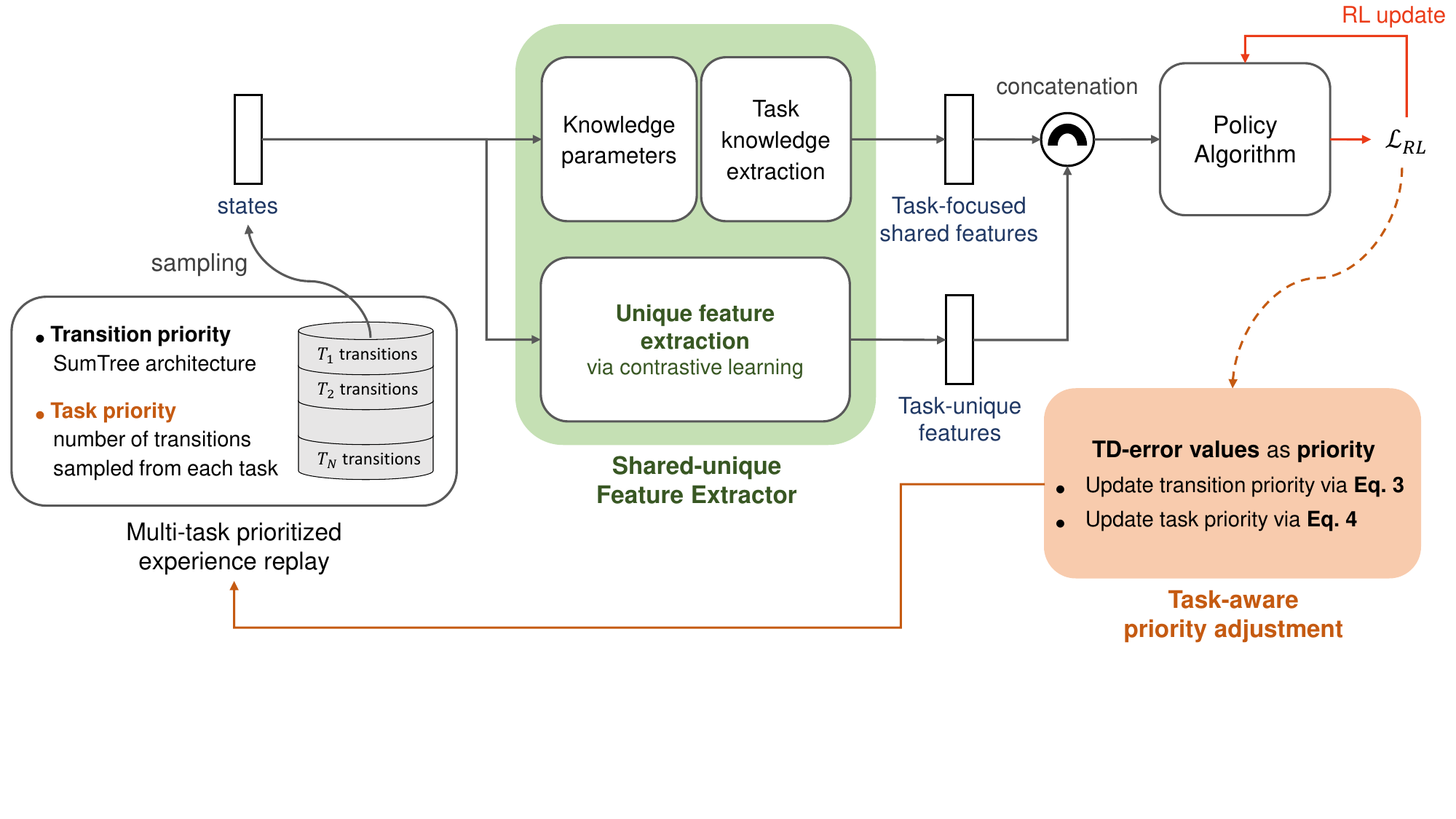}
    \vskip -0.65in
    \caption{Architecture of STARS. Our STARS consists of two main components: a share-unique feature encoder and a task-aware sampling strategy. With these two components, STARS enhances knowledge synergy between tasks and focus on tasks with poor learning outcomes (we refer to the overview paragraph in Section \ref{sec:method} for more details).}
    \label{fig:architecture}
    \vskip -0.05in
\end{figure}

\paragraph{Overview} We propose STARS, a novel MTRL method aimed at addressing the performance imbalance issue. We present the architecture and the pseudocode of STARS in Figure \ref{fig:architecture} and Algorithm \ref{alg:training}, respectively. In each iteration, STARS first collects transitions from all tasks and stores them in a multi-task prioritized experience replay buffer. Then, it samples the transitions from the buffer, considering each task's and each transition's priority. Next, STARS utilizes a shared-unique feature extractor (Section \ref{met:arc}) to extract task-focused shared features and task-unique features. The policy takes as input the concatenated features for RL training. Finally, STARS uses the TD-errors of sampled transitions to update the priorities (Section \ref{met:mtper}). Through these processes, STARS enhances the knowledge synergy between tasks and focuses on tasks with poor learning outcomes. 

\subsection{Shared-unique Feature Extractor}
\label{met:arc}

With the sampled transitions, we develop a shared-unique feature extractor to extract two informative features from each state: task-focused shared features and task-unique features. We explain why these features are required and how to obtain them now.

\textbf{Task-focused shared features:} The features encompass the essential knowledge or skills required for the current task, typically acquired from a shared knowledge pool. To achieve this, we adopt the architecture from PaCo \cite{sun2022paco}. 

Specifically, a shared set of knowledge parameters $\Phi \in \mathbb{R}^{P \times K}$ (referred to as the parameter set in \cite{sun2022paco}) captures $K$ knowledge, each characterized by $P$ learnable parameters. Subsequently, a series of transform weights (termed compositional vectors in \cite{sun2022paco}), denoted as $\mathrm{W} = [\mathrm{w}_{1}, \mathrm{w}_{j}, ..., \mathrm{w}_{N}]$, where $\mathrm{w}_{j} \in \mathbb{R}^{K}$, is employed to retrieve the relevant knowledge for each task from the shared $\Phi$. For a particular task $T_{j}$, the final parameters are computed as $\Omega_{j} = \Phi\mathrm{w}_{j}$, which are then utilized to extract task-focused shared features $f^{s}_{j}$  conditioned on the input states. During gradient backpropagation, the learning of shared $\Phi$ is influenced  by data from all tasks, while only data from task $T_{j}$ contributes to updating the corresponding transform weight $w_{j}$.

Now a question may arise: why do we need task-unique features if we already have task-focused shared features? The main reason is that this knowledge pool does not guarantee coverage of all the required knowledge or skills, especially considering the number and diversity of tasks. Additionally, it may be dominated by the knowledge required for the most similar tasks, as discussed in \cite{sun2022paco}. Therefore, we introduce task-unique features to retain the unique details for each task.

\textbf{Task-unique features: } The features aim to capture the detailed, unique knowledge for each task to provide more granular guidance. To achieve this, we construct several fully-connected layers to extract the task-unique features $f^{u}_{j}$ using states from each task $T_{j}$ as inputs. Then, except for being updated in the RL training, the features $f^{u}_{j}$ are also guided by the triplet loss \cite{Schroff15}, as formulated by

\begin{equation}
\label{equ:tri}
    \mathcal{L}_{tri} = \frac{1}{M} \sum_{i=1}^{M} \mathrm{max}(0, m + d(f^{u}_{i, a}, f^{u}_{i, p}) - d(f^{u}_{i, a}, f^{u}_{i, n})).
\end{equation}

Here, $M$ is the number of sampled pairs. Each pair contains three task-unique features, where the anchor $f^{u}_{a}$ and positive sample $f^{u}_{p}$ are from the same task $T_{j}$, and the negative sample $f^{u}_{n}$ is from another task $T_{\neq j}$. Moreover, $m$ is the margin in the original triplet loss, and we use Euclidean distance for the distance function $d(\cdot , \cdot)$. The task-unique features guided by $L_{tri}$ are encouraged to capture details in the current task, thus enhancing the policy to have a comprehensive understanding.

For each state, the policy extracts the two aforementioned features and concatenates them for the subsequent RL training. In Section \ref{sec:result}, we demonstrate that the policy learned with both features can achieve superior performance and effectively address the issue of performance imbalance.

\subsection{Task-aware Prioritized Sampling}
\label{met:mtper}
As mentioned in the Overview, the policy collected and stored the transitions in the replay buffer for subsequent RL learning. However, previous MTRL works usually use a basic experience replay or a single prioritized experience replay (PER) \cite{schaul2015prioritized} for transitions from all tasks, neglecting the characteristics of the MTRL task. Consequently, we developed a multi-task variant of PER in our STARS, which includes two levels of priorities: transition priority and task priority.

\textbf{Transition priority:} We use the temporal difference error (TD-error) value as the transition priority, so that transitions causing larger TD-errors will have higher priority. This setting can be applied to any RL algorithm with Q-networks. Since SAC is our base policy, we define the priority of an arbitrary transition tuple ($s, a, r, s^{'}$) as the average TD-errors from the two Q-networks in SAC: 

\begin{equation}
\label{equ:priority}
   \mathrm{priority} = |\delta| + \epsilon, \: \mathrm{where} \: |\delta| = \frac{1}{2} \sum_{i=1}^2 | r + \gamma V_{\theta_{tar}} (s') - Q_{\theta, i}(s,a)|.
\end{equation}

Here, $V_{\theta_{tar}} (s')$ is calculated from the target Q-network, and $\epsilon$ is a small positive constant to prevent no priority. For all $N$ tasks, we store transitions from each task to the corresponding PER. Moreover, if a transition is newly added to the PER, it will be set with the highest priority. Once a transition in the PER is sampled for RL training, the old priority will be directly replaced by the new value. Finally, we follow the steps in \cite{schaul2015prioritized} to calculate the probabilities and importance sampling weights when sampling transitions from the PER of a task.

\textbf{Task priority:} As mentioned above, each task contains its own PER, which can ensure that TD-error between different tasks will not interfere with each other, causing some tasks to have no transition sampled at all. Next, although we can allocate training resources evenly to all tasks, we would like to focus on those tasks that currently have poor learning outcomes. Therefore, we further consider each task's priority. Specifically, let $B$ be the total number of transitions that can be sampled. The number of transitions $B_{j}$ that can be sampled from each task $T_{j}$ is calculated as:

\begin{equation}
\label{equ:k_i}
    B_{j} = \mathrm{clip} ( B \cdot \frac{\mathrm{sum}(\mathrm{PER_{j}})}{\sum_{i=1}^{N} \mathrm{sum}(\mathrm{PER_{i}})}, B_{min}, B_{max}).
\end{equation}

Here, $\mathrm{sum}(\cdot)$ is a function that accumulates the priorities of all transitions in a PER. The $\mathrm{clip}$ function aims to balance the trade-off between even allocation and task priority, with $B_{min}$ and $B_{max}$ representing the minimum and maximum number of transition that can be sampled, respectively. With this multi-task PER design, STARS can dynamically adjust task priorities and reinforce tasks with poor learning outcomes during the learning process.

\begin{algorithm}[!t]
\caption{STARS pseudocode}\label{alg:cap}
\label{alg:training}
\textbf{Input:} parameters $\theta$ (all parameters in the feature extractor and SAC policy) \\
\textbf{Input:} batchsize $B$, task lists $\mathcal{T} =[T_{1}, T_{j}, ..., T_{N}]$ \\
\textbf{Output:} learned parameters $\theta$ 
\begin{algorithmic}[1]
\State Randomly initialize parameters $\theta$
\State Create and initialize all $\mathrm{PER}$s (replay buffers)
\While {termination condition is not satisfied}
    
    \State Collect transitions using current policy and update PERs if needed
    \State $\mathrm{batch} \leftarrow [ \: ]$ 
    \State Allocate training resources $B$ via \textbf{Eq. \ref{equ:k_i}}
    \For{all task $T_{j}$ in $\mathcal{T}$}
        \State Sample $B_{j}$ transitions from $\mathrm{PER}_{j}$
        \State  Add $B_{j}$ transitions into $\mathrm{batch}$
    \EndFor
    \State $\mathcal{L}_{RL}, \mathcal{L}_{tri} \leftarrow$ STARS training on $\mathrm{batch}$ 
    \State Backpropagate gradients from $\mathcal{L}_{RL}, \mathcal{L}_{tri}$ to update parameters $\theta$
    \For{each transition in $\mathrm{batch}$}
    \State Compute new priority via \textbf{Eq. \ref{equ:priority}} (using TD-error in $\mathcal{L}_{RL}$)
    \State Update its corresponding $\mathrm{PER}$ with the new priority
    \EndFor
\EndWhile
\end{algorithmic}
\end{algorithm}

\section{Experiments}
\label{sec:result}
We conduct experiments to verify two questions: (1) Does our STARS, which considers both shared and unique features along with task-aware sampling, effectively address the performance imbalance issue? (2) Does STARS outperform state-of-the-art methods in MTRL?

\subsection{Experimental Setting}
\paragraph{Evaluation Tasks} We evaluate STARS and baselines through experiments on the Meta-World benchmark \cite{yu2020meta}, a popular platform for meta-RL and MTRL studies. This benchmark provides a diverse set of robot manipulation tasks to assess methods' effectiveness in complex and realistic environments. To better align with real-world scenarios, we adopt modifications from previous works \cite{yang2020multi, sun2022paco} for the benchmark, randomly initializing the poses of objects in the environment. Evaluations are conducted on both the MT-10 and MT-50 tracks, which contain 10 and 50 tasks, respectively.

\paragraph{Compared Baselines} Our STARS is compared with six representative baselines: \textbf{(1)} \textbf{Oracle}: A SAC policy  \cite{haarnoja2018soft} trained individually for each task. \textbf{(2)} Multi-task SAC \textbf{(MT-SAC)} \cite{yu2020meta}: A unified SAC policy with an additional one-hot task encoding as input. \textbf{(3)} Projecting Conflicting Gradients \textbf{(PCGrad)} \cite{yu2020gradient}: A method projecting gradients to eliminate gradient conflicts between different tasks, thereby mitigating negative influence. \textbf{(4)} \textbf{Soft Modularization} \cite{yang2020multi}: A method developing a routing network that generates routes for different tasks in a modularized manner, allowing the policy network to adapt to each task. \textbf{(5)} Contextual Attention-based REpresentation \textbf{(CARE)} \cite{sodhani2021multi}: A method utilizing the embedding of each task's description to present the task information, allowing advanced context-aware learning. \textbf{(6)} Parameter-Compositional Multi-Task Reinforcement Learning \textbf{(PaCo)} \cite{sun2022paco}: A method leveraging a shared parameter set and task-specific vectors to learn which knowledge is required for each task.

\paragraph{Evaluation Metrics $\&$ Details} Following previous MTRL works \cite{sodhani2021multi, sun2022paco}, we report the average success rate (SR) across multiple experiment runs (referred to as \textbf{across runs} for short) to ensure a fair comparison. Additionally, we report the average SR \textbf{across tasks} in some experiments to validate the methods' improvements on the performance imbalance issue. Specifically, we conduct different experiment runs with 10 random seeds. In each experiment run, the maximum environment interaction is set to 2 million steps per task (20 million and 100 million in total for MT-10 and MT-50, respectively). Moreover, every 250k steps, each method is tested in a test environment for 50 episodes to compute its success rate at that time point. 

The difference between the two average SRs lies in the order of calculating the average. For the first one (across runs), the SR at each time point is first averaged across tasks and then averaged across runs. The best score from theses time points will be used for comparison. For the second one (across tasks), the SR at each time point is first averaged across runs and then averaged across tasks. We report the average SRs at each time point to demonstrate the performance imbalance issue.

\subsection{Experimental Analysis}

\begin{table*}[ht]
\centering
\caption{Performance comparison on the MT-10 track. Average SR [$\uparrow$] (across runs) is reported. Our STARS statistically outperforms the the Oracle and previous MTRL methods.}
\label{tab:result}
\begin{tabular}{lc}
\toprule
\multirow{2}{*}{Methods} & average SR(\%) \\
& mean $\pm$ std \\
\midrule
Oracle & 80.6 $\pm$ 4.2 \\
\midrule
MT-SAC~\citep{yu2020meta} & 56.7 $\pm$ 7.5\\
PCGrad~\citep{yu2020gradient} & 59.4 $\pm$ 8.9 \\
Soft Modularization~\citep{yang2020multi} & 65.8 $\pm$ 4.5\\
CARE~\citep{sodhani2021multi} & 78.2 $\pm$ 5.8\\
PaCo~\citep{sun2022paco} & 83.1 $\pm$ 4.6\\
\midrule
STARS (Ours) & \textbf{88.5 $\pm$ 5.3}\\
\bottomrule
\end{tabular}
\end{table*}

\paragraph{Main Experiment} Our main experiment evaluates the overall performance of each method, consistent with the main research goals of previous MTRL works. In Table \ref{tab:result}, we summarize the best average SR of each method achieved during training in the MT-10 track, with the standard deviation (STD) calculated across multiple experiment runs. From the results, we have the following findings: 

\textbf{Finding $\#$1:} When comparing STARS with previous MTRL SOTA methods, it is evident that our STARS outperforms them with an improvement of $6.5\%$. Considering that the performance might overlap after including the standard deviation, we further conducted a statistical test and concluded that our performance is statistically significantly better than the previous method. The statistical test process is detailed in Appendix \ref{sec:statitics}. 

\textbf{Finding $\#$2:} The Oracle method, which trains a SAC separately for each task and averages their performance, is generally considered the best way to focus on a single task to achieve the highest performance on that task. However, surprisingly, the performance of our STARS is significantly higher than that of the Oracle method. This indicates that STARS effectively leverages cross-task knowledge synergy to achieve better performance than training on each task independently.

Regarding the MT-50 track, we have compiled the results in Appendix \ref{sec:mt50}. Our findings show that while STARS still achieves the best results, the performance gap with previous methods is not as significant. Effectively utilizing the shared and unique knowledge between tasks and adjusting priorities among tasks are our future research goals.

\begin{table*}[ht]
\centering
\caption{Performance Difference across tasks. Average SR [$\uparrow$] (across tasks) is reported. This table illustrates the actual values for the performance imbalance issue depicted in Figure \ref{fig:issue}. The performances of previous MTRL methods are significantly unbalanced. In contrast, our STARS method not only achieves the best performance but also exhibits the highest stability.}
\label{tab:task_average}
\begin{adjustbox}{width={\textwidth},totalheight={\textheight},keepaspectratio}
\begin{tabular}{lcccccccc}
\toprule
Steps (10k) & 25 & 50 & 75 & 100 & 125 & 150 & 175 & 200\\
\midrule
CARE~\citep{sodhani2021multi} & 31.5 $\pm$ 37.3 & 39.1 $\pm$ 36.8 & 55.0 $\pm$ 30.4 & 52.1 $\pm$ 39.7 & 60.6 $\pm$ 20.2 & 75.8 $\pm$ 19.1 & 58.9 $\pm$ 28.2 & 78.3 $\pm$ 22.9 \\
PaCo~\citep{sun2022paco} & 57.0 $\pm$ 38.9 & 70.3 $\pm$ 44.0 & 73.8 $\pm$ 38.1 & 74.3 $\pm$ 39.7 & 77.1 $\pm$ 32.7 & 83.0 $\pm$ 30.6 & 80.6 $\pm$ 31.4 & 81.7 $\pm$ 30.0 \\
\midrule
STARS (Ours) & \textbf{58.8 $\pm$ 10.8} & \textbf{70.5 $\pm$ 2.4} & \textbf{81.5 $\pm$ 6.6} & \textbf{79.2 $\pm$ 6.7} & \textbf{80.2 $\pm$ 5.7} & \textbf{87.0 $\pm$ 3.7} & \textbf{85.0 $\pm$ 8.4} & \textbf{85.8 $\pm$ 6.0} \\
\bottomrule
\end{tabular}
\end{adjustbox}
\end{table*}

\paragraph{Performance Imbalance Issue}
\label{subsec:task-imbalance}

As discussed in Section \ref{sec:intro}, previous MTRL methods strive to design different architectures to utilize knowledge from different tasks and switch policy modes or extract required knowledge when solving specific tasks. However, they did not take into account the knowledge that can be shared between tasks and the details that each requires special attention at the same time, resulting in the serious performance imbalance issue. In other words, they may be competitive on average performance but have extremely poor learning outcomes on certain tasks.

Therefore, we developed STARS, an MTRL method that considers both shared knowledge and unique details between tasks. In addition, we designed a transition sampling strategy for MTRL tasks, TaPS, which can dynamically allocate more training resources to tasks with poor current learning outcomes while ensuring a balance between tasks. As a result, our STARS not only achieves the best performance on the challenging MT-10 track, but also has the smallest performance standard deviation across tasks, showing its high stability. The result is illustrated in Table \ref{tab:task_average}.

\subsection{Ablation Study}
\label{sec:ablation}

\newcommand{\cmark}{\ding{51}}%
\newcommand{\xmark}{\ding{55}}%
\begin{table*}[!ht]
    \centering
    \caption{Component contributions of STARS. Average SR [$\uparrow$] (across runs) is reported. Either of the two components helps STARS achieve brtter results, especially when STARS integrates both of them.}
    \begin{tabular}{ccc}
    \toprule
    \multicolumn{2}{c}{Methods} & average SR(\%) \\
     TaPS & Unique Features & mean $\pm$ std \\
    \midrule
    \xmark & \xmark & 83.1 $\pm$ 4.6 \\
    \cmark & \xmark & 86.4 $\pm$ 3.8 \\
    \xmark & \cmark & 85.7 $\pm$ 6.2 \\
    \cmark & \cmark & 88.5 $\pm$ 5.3 \\
    \bottomrule
    \end{tabular}
    
    \label{tab:abl}
\end{table*}

\paragraph{Component Contributions} In this paragraph, we conducted an ablation study to verify the performance contributed by our designed components. Our STARS has two main components: a share-unique feature extractor and task-aware prioritized sampling (TaPS). The first component aims to capture both shared and task-unique features for better knowledge synergy across tasks. The later component dynamically adjusts the allocation of training resources based on the learning outcomes for each task. From the results in Table \ref{tab:abl}, we found that the model integrated with either component shows improved performance. Moreover, the model with both components achieves the highest scores, demonstrating that our motivation and design are effective and well-supported.

\paragraph{Transition Sampling Strategies}

\begin{table*}[!ht]
\centering
\caption{Performance comparison of various sampling strategies. Average SR [$\uparrow$] (across runs) is reported. STARS with our task-aware prioritized sampling (TaPS) achieves the highest performance. }
\label{tab:sampling}
\begin{tabular}{ccccc}
\toprule
 & random & task-equally random & PER & TaPS (Ours) \\
\midrule
average SR & 85.1$\pm$ 4.5 &  86.2 $\pm$ 4.3 &  83.3 $\pm$ 5.2 & \textbf{88.5 $\pm$ 5.3} \\
\bottomrule
\end{tabular}
\end{table*}

To demonstrate the necessity of designing novel sampling strategies tailored to the characteristics of MTRL tasks, this ablation study compares our newly proposed TaPS sampling strategy with three other widely-used sampling strategies in the RL literature: random sampling, task-equally random sampling, and prioritized experience replay (PER) \cite{schaul2015prioritized}. We briefly introduce them and present the challenges encountered when applying them to MTRL tasks below.

 \textbf{Random sampling}: this strategy uniformly and randomly selects transitions from all tasks, resulting in an inconsistent number of sampled transitions from different tasks. It is used by most previous MTRL studies. \textbf{Task-equally random sampling}: this strategy distributes training resources (i.e., the total number of sampled transitions) evenly among all tasks, and then randomly samples from the replay buffer of each task. Although it allocates resources fairly, this strategy overlooks the priority of transitions within the same task and the task priority among different tasks. \textbf{PER} \cite{schaul2015prioritized}: this strategy assigns a priority to each transition and uses these priorities to weight the likelihood of being sampled. However, using a single PER in the MTRL task may cause the transitions of a certain task to be not sampled because the priority value range of different tasks may vary, resulting in a significant deviation in the distribution of sampled transitions.

We integrate STARS with these sampling strategies and present the results in Table \ref{tab:sampling}, which echoes our views. First, the results of random sampling and task-equally random sampling are similar. This is because the policy collects the same number of transitions in each task and adds them into the replay buffer. Therefore, using random sampling will have similar expectation as using task-equally random sampling in this setting. However, ensuring an even distribution of resources every time (i.e., task-equally random) will still achieve more stable results in practice. Next, using a single PER yielded the worst results, which is expected. Because the distribution of sampled transitions is highly biased, the policy will be overly biased for certain tasks when learning. In contrast, our TaPS simultaneously considers the resource balance between tasks and dynamically adjusts the tasks that currently require more attention, thus obtaining the best results.


\begin{table*}[ht]
    \centering
    \caption{Performance comparison between baselines w/ and w/o TaPS. Average SR [$\uparrow$] (across runs) is reported. Most baselines benefit from our TaPS and achieve better results.}
    \label{tab:mt}
    \begin{tabular}{lccccc}
        \toprule
         & MT-SAC  &  PCGrad & Soft Modularization & CARE & PaCo \\
        \midrule
         w/o TaPS  & 56.7 $\pm$ 7.5 & 59.4 $\pm$ 8.9 & \textbf{65.8 $\pm$ 4.5} & 78.2 $\pm$ 5.8 & 83.1 $\pm$ 4.6 \\
        w/ TaPS   & \textbf{60.1 $\pm$ 8.3} &  \textbf{61.3 $\pm$ 10.1} & 60.1 $\pm$ 4.9 & \textbf{79.1 $\pm$ 6.9} & \textbf{86.4 $\pm$ 3.8}\\
        \bottomrule
    \end{tabular}
\end{table*}

\paragraph{Baselines with Task-aware Prioritized Sampling (TaPS)} Since our developed TaPS is a transition sampling strategy that can be easily combined with baselines, Table \ref{tab:mt} illustrates the performance of baselines with and without our TaPS. This allows us to verify whether TaPS is highly general and can cooperate with multiple methods. The results show that most of the baselines integrated with our TaPS achieve better performance, supporting our claims. However, the reason why Soft Modularization with our TaPS results in a worse outcome requires further investigation.

\begin{wrapfigure}{r}{0.64\textwidth}
  \centering
  \includegraphics[width=0.64\textwidth]{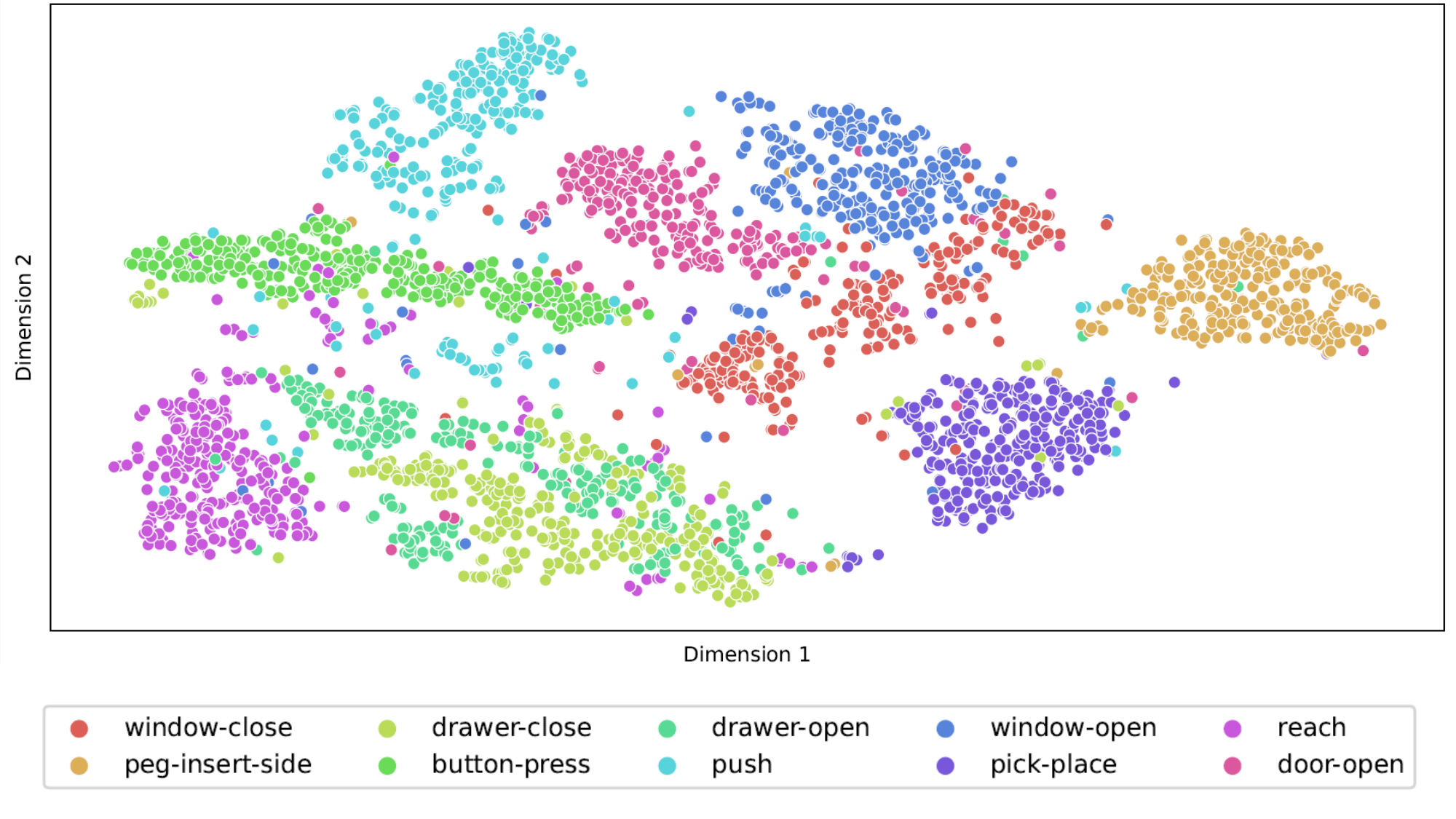}
  \caption{t-SNE visualization of the learned unique features of our STARS on the MT-10 track.}
  \label{fig:mt10_tsne}

  \vskip -0.15in
\end{wrapfigure}

\paragraph{Embedding Visualizations} Since we newly introduced task-unique features in our STARS, we performed t-SNE analysis \cite{Laurens_tSNE} on these unique features and presented the results in Figures \ref{fig:mt10_tsne} to further verify their effectiveness. In the figure, points with the same color represent unique features extracted from the same task. 

From Figure \ref{fig:mt10_tsne}, it is evident that the contrastive objective we use allows the unique features learned to retain the unique properties of tasks on the MT-10 track, making features from the same task more similar. However, from the MT-50 track results shown in Figure \ref{fig:mt50_tsne} in the Appendix, it is difficult to preserve the unique details of each task, resulting in the inability to effectively distinguish these unique features. This also reflects in the smaller performance gain our STARS achieved on the MT-50 track . How to obtain better knowledge synergy with more tasks is our future research direction.
\section{Conclusion}
\label{sec:conclusion}
In this paper, we point out the serious performance imbalance issue suffered by previous MTRL methods and analyze the reasons why it occurs. Previous methods were unable to (1) simultaneously consider the knowledge that can be shared between tasks and the details that require special attention, and (2) adjust the priority of each task during the learning process. Therefore, we propose STARS, a new MTRL method that uses a share-unique feature encoder to solve (1) and alleviates (2) by leveraging a task-aware prioritized sampling strategy to dynamically adjust the training resources allocated to each task. We compare STARS with other SOTA MTRL methods on the Meta-World benchmark and achieve statistically superior results (6.5$\%$ improvement) on the MT-10 track and better results on the MT-50 track. In addition, we conducted a number of ablation studies and provided embedding visualizations to demonstrate that our STARS effectively solves the issue of performance imbalance. Finally, we responsibly present the limitations of our work in Appendix \ref{sec:limitation}.

{
\small
\bibliographystyle{unsrtnat}
\bibliography{main}
}

\newpage
\appendix
\section{More Details on Performance Imbalance Issue}
\label{sec:compare}

\begin{figure}[h]
    \centering
    \begin{subfigure}[b]{0.94\textwidth}
        \centering
        \includegraphics[width=0.94\textwidth]{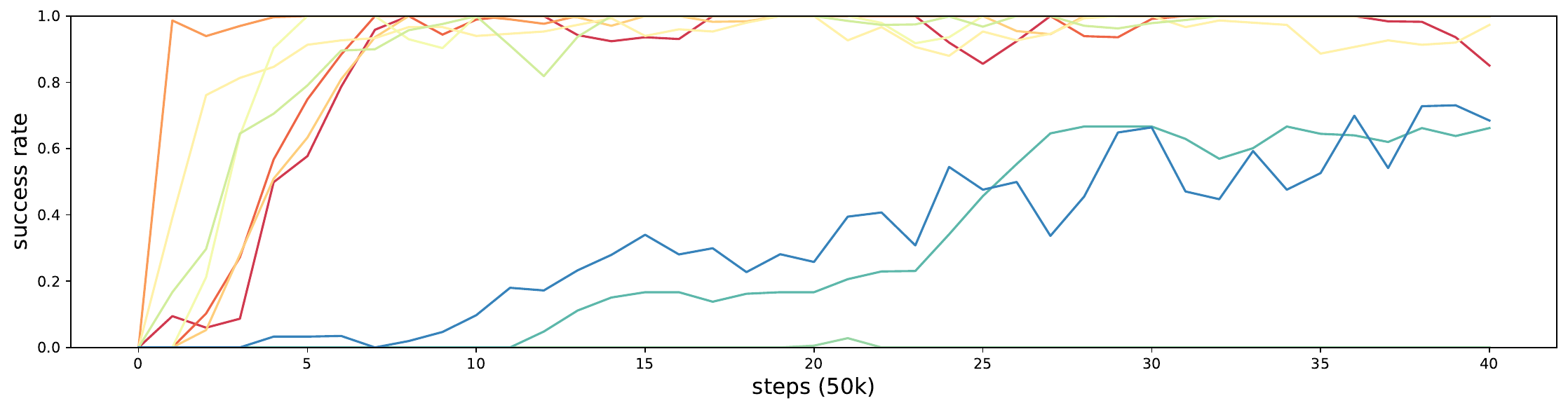}
        \caption{PaCo}
        \label{fig:paco}
    \end{subfigure}

    \begin{subfigure}[b]{0.94\textwidth}
        \centering
        \includegraphics[width=0.94\textwidth]{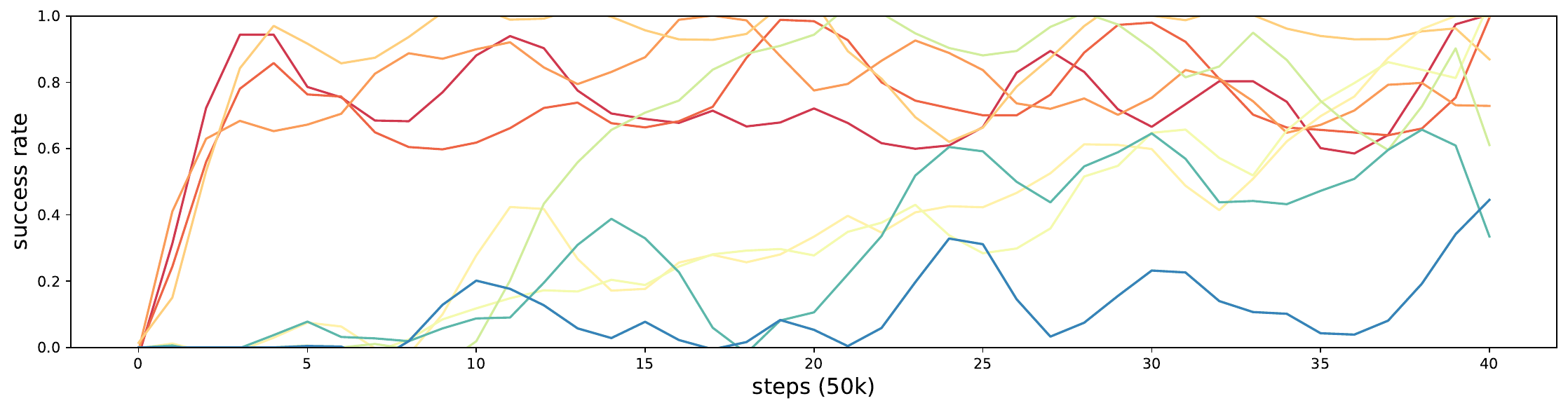}
        \caption{CARE}
        \label{fig:care}
    \end{subfigure}

    \begin{subfigure}[b]{0.94\textwidth}
        \centering
        \includegraphics[width=0.94\textwidth]{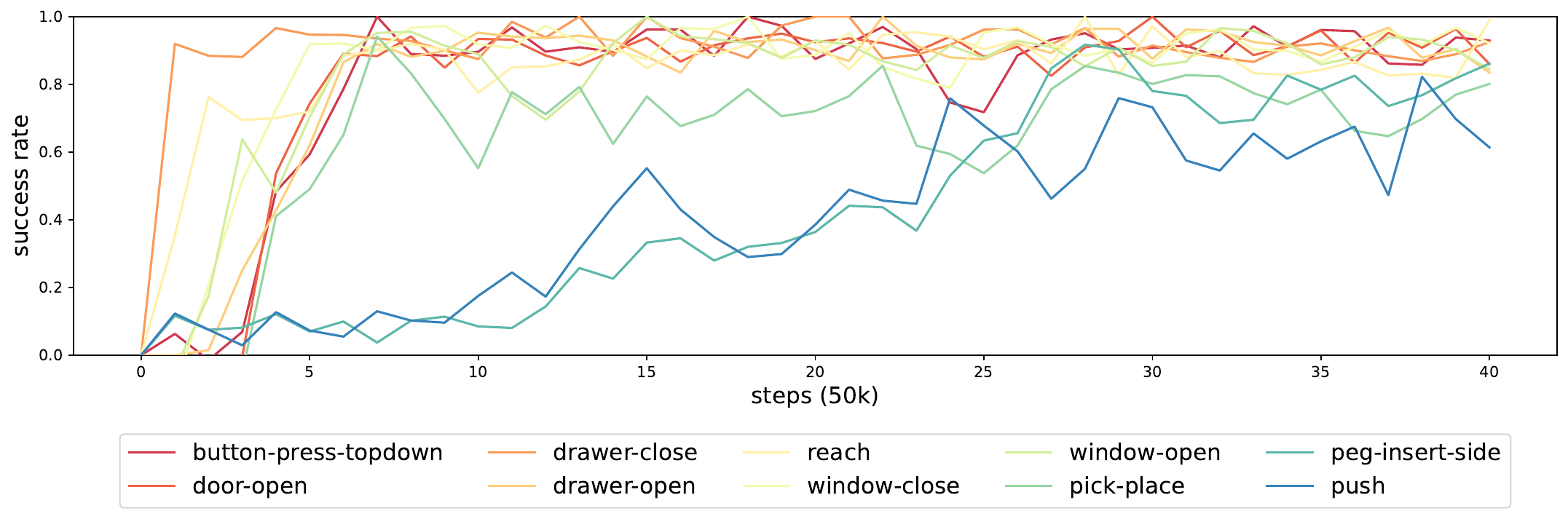}
        \caption{STARS (Ours)}
        \label{fig:ours}
    \end{subfigure}
    
    \caption{Performance Imbalance Issue (per task). We have presented the issue with the average success rate (SR) across tasks in Figure \ref{fig:issue} and Table \ref{tab:result} of the main paper. Here, we provide the average SR (already averaged over 10 runs) for each task. Our STARS method effectively focuses on the more difficult tasks and demonstrates the best performance stability among the compared methods.}

    \label{fig:issue_per_task}
\end{figure}
\begin{table*}[ht]
\centering
\caption{Performance comparison on the selected \textit{peg-insert-side} and \textit{pick-place} tasks in the MT-10 track. Average SR [$\uparrow$] (across runs) is reported. Most methods perform poorly on both tasks when solving the MT-10 track. However, our STARS method can effectively learn these two tasks and consistently achieves better results than other methods during the learning process.}
\label{tab:peg-and-pick}
\begin{adjustbox}{width={\textwidth},totalheight={\textheight},keepaspectratio}
\begin{tabular}{lcccccccc}
\toprule
Steps (10k) & 25 & 50 & 75 & 100 & 125 & 150 & 175 & 200\\
\midrule
\midrule
\multirow{2}{*}{CARE~\citep{sodhani2021multi}} 
& \textbf{11.6 $\pm$ 5.8} & \textbf{19.6 $\pm$ 20.1} & \textbf{52.0 $\pm$ 39.8} & 0.0 $\pm$ 0.0 & 60.0 $\pm$ 45.0 & 72.4 $\pm$ 41.1 & 32.0 $\pm$ 42.3 & 35.2 $\pm$ 24.9 \\
& 0.0 $\pm$ 0.0 & 11.2 $\pm$ 0.0 & 16.4 $\pm$ 0.0 & 15.6 $\pm$ 1.2 & 48.4 $\pm$ 0.0 & 35.6 $\pm$ 0.0 & 3.6 $\pm$ 0.0 & 44.0 $\pm$ 0.0 \\
\midrule
\multirow{2}{*}{PaCo~\citep{sun2022paco}} & 0.0 $\pm$ 0.0 & 0.0 $\pm$ 0.0 & 16.7 $\pm$ 37.3 & 16.7 $\pm$ 37.3 & 45.7 $\pm$ 42.0 & 66.7 $\pm$ 47.1 & 64.5 $\pm$ 45.8 & 66.2 $\pm$ 46.8 \\
& 0.0 $\pm$ 0.0 & 0.0 $\pm$ 0.0 & 0.0 $\pm$ 0.0 & 0.5 $\pm$ 1.2 & 0.0 $\pm$ 0.0 & 0.0 $\pm$ 0.0 & 0.0 $\pm$ 0.0 & 0.0 $\pm$ 0.0 \\
\midrule
\multirow{2}{*}{Ours} & 7.0 $\pm$ 7.1 & 8.5 $\pm$ 8.5 & 33.3 $\pm$ 39.2 & \textbf{36.4 $\pm$ 46.8} & \textbf{63.4 $\pm$ 45.5} & \textbf{78.0 $\pm$ 45.3} & \textbf{78.4 $\pm$ 45.8} & \textbf{86.1 $\pm$ 44.0} \\
& \textbf{49.0 $\pm$ 34.1} & \textbf{55.2 $\pm$ 6.5} & \textbf{76.4 $\pm$ 20.1} & \textbf{72.1 $\pm$ 21.8} & \textbf{53.8 $\pm$ 7.2} & \textbf{80.1 $\pm$ 16.2} & \textbf{78.4 $\pm$ 13.7} & \textbf{80.1 $\pm$ 16.2}  \\
\bottomrule
\end{tabular}
\end{adjustbox}
\end{table*}

This section provides more details and analysis on the performance imbalance issue. Figure \ref{fig:issue_per_task} shows the learning curves of the three methods on 10 tasks in the MT-10 track. It can be observed that two tasks, \textit{peg-insert-side} and \textit{pick-place}, are relatively difficult, and the knowledge and skills they require are different from those needed for other task. Moreover, the previous MTRL methods ignored task priority adjustment during the learning process, and the knowledge it focused on was dominated by other tasks, resulting in poor learning outcomes for these two tasks, as shown in Table \ref{tab:peg-and-pick}. 

In contrast, because of the novel design of the share-unique feature encoder and task-aware prioritized sampling strategy, our STARS method can monitor its learning outcomes on these two tasks and allocate more attention to them during training, allowing STARS to achieve the best overall performance while enhancing its performance on these two tasks.

\section{Experiment Results on the MT-50 Track}
\label{sec:mt50}
\begin{table*}[!ht]
\centering
\caption{Performance comparison on the MT-50 track. Average SR [$\uparrow$] (across runs) is reported. Our method achieved the best performance overall; however, the differences compared to other methods were not statistically significant.}
\label{tab:mt50}
\begin{tabular}{lc}
\toprule
\multirow{2}{*}{Methods} & success rate(\%) \\
& (mean + std) \\
\midrule
MT-SAC~\citep{yu2020meta} & 38 $\pm$ 2.1\\
PCGrad~\citep{yu2020gradient} & 45 $\pm$ 3.8 \\
Soft Modularization~\citep{yang2020multi} & 48 $\pm$ 2.6\\
CARE~\citep{sodhani2021multi} & 52.3 $\pm$ 3.7\\
PaCo~\citep{sun2022paco} & 55.1 $\pm$ 2.4\\
\midrule
Ours & \textbf{56.2 $\pm$ 3.1}\\
\bottomrule
\end{tabular}
\end{table*}

\begin{figure}[!ht]
    \centering
    \includegraphics[width=0.9\linewidth]{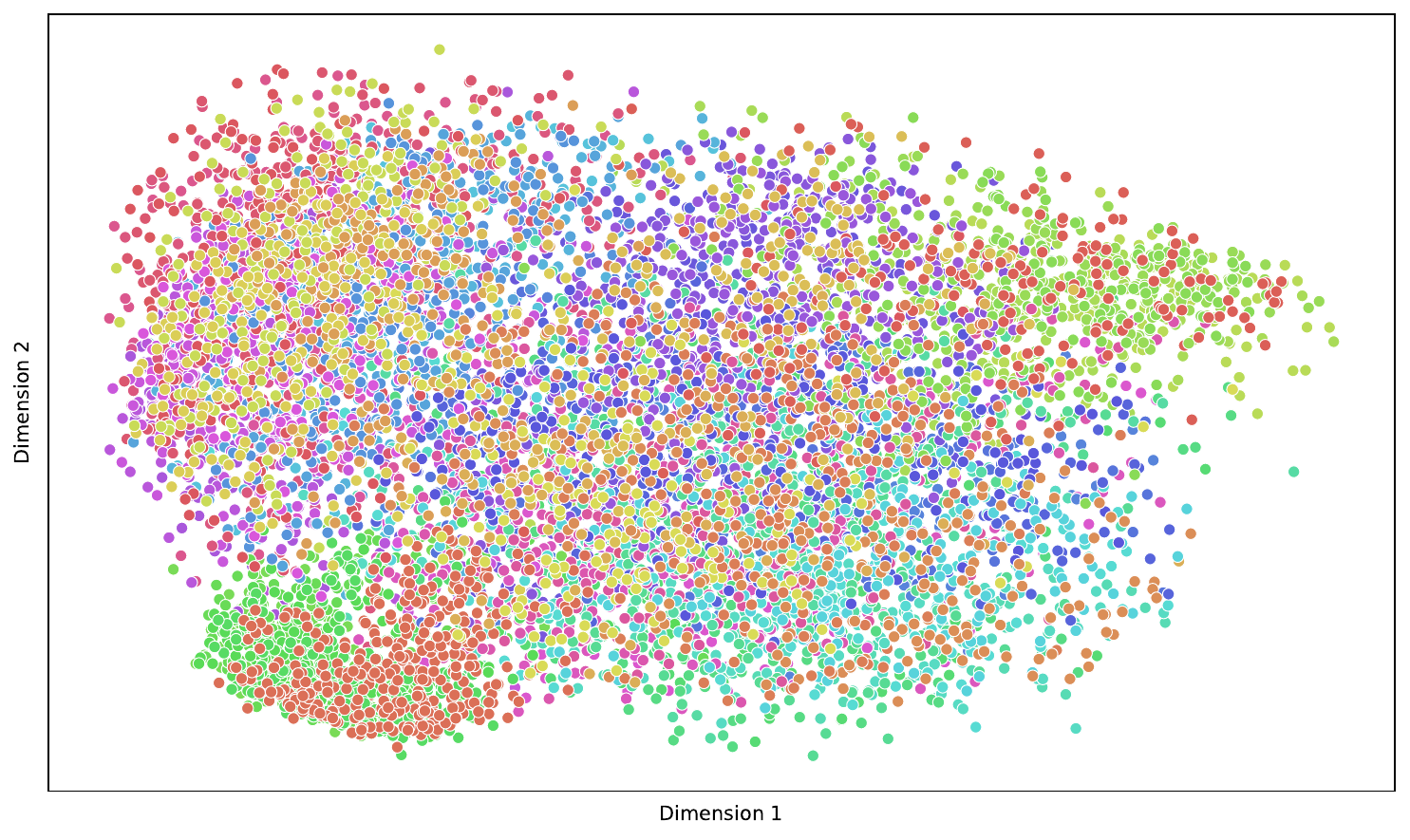}
    \caption{t-SNE visualization of the learned unique features of our STARS on the MT-50 track.}
    \label{fig:mt50_tsne}
\end{figure}

For a more comprehensive comparison, we also evaluated our STARS method against other MTRL methods on the MT-50 track, and the results are summarized in Table \ref{tab:mt50}. Since the MT-50 track contains 50 diverse robot manipulation tasks, more knowledge and skills need to be extracted and analyzed, making monitoring during the learning process more challenging. Nevertheless, our STARS method still achieves the best results among all methods, indicating out that the two novel components we designed are indeed helpful. However, the performance gap with other methods is not statistically significant. This can also be observed from the t-SNE analysis of the unique features of the MT-50 track, as shown in Figure \ref{fig:mt50_tsne}. In this figure, unique features from the same task cannot be effectively clustered together, and unique features from different tasks cannot be effectively separated.

Effectively obtaining better knowledge synergy when facing more tasks and focusing more accurately on tasks that need to be strengthened is our future research direction. Possible directions include designing more innovative architectures, developing different contrastive objectives to extract unique features, or using more advanced strategies to allocate training resources. We will continue to conduct experiments in these directions.

\section{Statistical Analysis of Performance Difference on the MT-10 Track}
\label{sec:statitics}
Considering that in the MT-10 track (Table \ref{tab:result}), the performance of our STARS method and the second-best PaCo may overlap after accounting for the standard deviation, we conducted a statistical analysis to confirm whether the performance obtained by our STARS method is statistically significantly superior to that of PaCo. Specifically, we performed a two-sample t-test. The t-test will compare the means of the two methods and determine if the difference is statistically significant.

Given:
\begin{itemize}
    \item Mean of our STARS (\(\bar{x}_1\)): 88.5
    \item Standard deviation of our STARS (\(s_1\)): 5.3
    \item Mean of PaCo (\(\bar{x}_2\)): 83.1
    \item Standard deviation of PaCo (\(s_2\)): 4.6
    \item Number of runs (\(n\)): 10
\end{itemize}

We use the following formula for the t-statistic in a two-sample t-test:
\[
t = \frac{\bar{x}_1 - \bar{x}_2}{\sqrt{\frac{s_1^2}{n} + \frac{s_2^2}{n}}}
\]

And the degrees of freedom for the test is calculated as:
\[
df = \frac{\left(\frac{s_1^2}{n} + \frac{s_2^2}{n}\right)^2}{\frac{\left(\frac{s_1^2}{n}\right)^2}{n-1} + \frac{\left(\frac{s_2^2}{n}\right)^2}{n-1}}
\]

First, we calculate the t-statistic:
\[
t = \frac{88.5 - 83.1}{\sqrt{\frac{5.3^2}{10} + \frac{4.6^2}{10}}} \approx 2.433
\]

Next, we calculate the degrees of freedom:
\[
df = \frac{\left(\frac{5.3^2}{10} + \frac{4.6^2}{10}\right)^2}{\frac{\left(\frac{5.3^2}{10}\right)^2}{9} + \frac{\left(\frac{4.6^2}{10}\right)^2}{9}} \approx 17.651
\]

Finally, we calculate the p-value using the t-distribution:
\[
p = 2 \times \text{Pr}(T > |t|) = 0.0258
\]

Since the p-value is 0.0258, which is less than the common significance level of 0.05, we conclude that there is a statistically significant difference between the performance of our STARS and PaCo. Therefore, \textbf{our STARS statistically outperforms PaCo on the MT-10 track}.

\section{Limitations}
\label{sec:limitation}

As mentioned in Appendix \ref{sec:mt50}, the main limitation of our STARS method is that the performance difference obtained when facing more tasks is not significant enough. This may be attributed to the limitations of the current architecture for extracting shared and unique features (such as the pre-defined number of the learnable knowledge in the knowledge pool), or the insufficient degree of dynamic adjustment of task priorities. We will continue to develop more novel architectures to extract shared and unique features to achieve better knowledge synergy. Additionally, we will employ more advanced sampling strategies to focus more accurately on tasks that require enhanced performance.


\end{document}